\documentclass[11pt]{article}
\usepackage{acl}
\usepackage{times}
\usepackage{latexsym}
\usepackage[T1]{fontenc}
\usepackage[utf8]{inputenc}
\usepackage{microtype}
\usepackage{inconsolata}

\usepackage{booktabs}
\usepackage{multirow}
\usepackage{graphicx}
\usepackage{amsmath}

\title{Seeing is Not Understanding: A Benchmark on Perception–Cognition Disparities in Large Language Models}

\usepackage{authblk}
\author[1]{Haokun Li}
\author[1]{Yazhou Zhang\textsuperscript{*}}
\author[2]{Jizhi Ding}
\author[3]{Qiuchi Li}
\author[1]{Peng Zhang\textsuperscript{*}}
\affil[1]{Tianjin University}
\affil[2]{Shandong Institute of Petroleum and Chemical Technology}
\affil[3]{Beijing Institute of Technology}
\begin{document}
\maketitle
\begin{abstract}
With the rapid advancement of Multimodal Large Language Models (MLLMs), they have demonstrated exceptional capabilities across a variety of vision-language tasks. However, current evaluation benchmarks predominantly focus on objective visual question answering or captioning, inadequately assessing the models' ability to understand complex and subjective human emotions. To bridge this gap, we introduce EmoBench-Reddit, a novel, hierarchical benchmark for multimodal emotion understanding. The dataset comprises 350 meticulously curated samples from the social media platform Reddit, each containing an image, associated user-provided text, and an emotion category (sad, humor, sarcasm, happy) confirmed by user flairs. We designed a hierarchical task framework that progresses from basic perception to advanced cognition, with each data point featuring six multiple-choice questions and one open-ended question of increasing difficulty. Perception tasks evaluate the model's ability to identify basic visual elements (e.g., colors, objects), while cognition tasks require scene reasoning, intent understanding, and deep empathy integrating textual context. We ensured annotation quality through a combination of AI assistance (Claude 4) and manual verification.We conducted a comprehensive evaluation of nine leading MLLMs, including GPT-5, Gemini-2.5-pro, and GPT-4o, on EmoBench-Reddit. 
\end{abstract}

\section{Introduction}

In recent years, Multimodal Large Language Models (MLLMs) have achieved breakthrough progress. Their ability to process and understand information from different modalities, such as text, images, and audio, has demonstrated powerful capabilities in domains like image captioning \cite{chen2015microsoft}, visual question answering \cite{antol2015vqa}, and cross-modal retrieval. The emergence of models like GPT-4V \cite{openai2023gpt4v} and Gemini \cite{google2023gemini} has further propelled the development of Artificial General Intelligence (AGI), enabling more natural and intelligent human-computer interactions.

However, the current MLLM evaluation landscape primarily focuses on the understanding and description of objective facts. Benchmarks such as MME \cite{fu2023mme} and MMBench \cite{liu2023mmbench} measure multimodal capabilities by assessing performance on tasks like OCR, object recognition, and spatial relationship judgment. While necessary, these evaluations largely overlook a more advanced and core aspect of intelligence: Emotional Intelligence (EI). EI—the ability to perceive, understand, manage, and utilize emotions—is key to achieving true human-machine empathy and harmonious interaction. A model incapable of understanding a user's subtle emotions, such as sarcasm or implicit sadness, will have its application scenarios severely limited.

To address the lack of depth in emotion understanding within current evaluation systems, we argue for a benchmark that can systematically and hierarchically assess the emotional intelligence of MLLMs. Such a benchmark should possess the following characteristics:
\begin{enumerate}
    \item \textbf{Data Authenticity:} Data should originate from real-world scenarios, not lab environments, to reflect the complexity and diversity of human emotional expression.
    \item \textbf{Task Hierarchy:} Evaluation tasks should progress from simple visual perception to complex cognitive reasoning, allowing for precise diagnosis of where a model's capabilities fall short.
    \item \textbf{Emotional Diversity:} It should cover a range of typical human emotions, especially those reliant on complex context and social knowledge, such as sarcasm.
    \item \textbf{Comprehensive Evaluation:} It should combine multiple-choice and open-ended questions to ensure both objective assessment and an evaluation of the accuracy and richness of generated content.
\end{enumerate}

Based on these considerations, we constructed \textbf{EmoBench-Reddit}, a hierarchical benchmark specifically designed to evaluate the emotional intelligence of MLLMs. We chose Reddit as our data source due to its vast repository of user-generated image-text content, where "flairs" are often used to tag a post's theme or emotion, providing a natural source of high-quality, labeled data. Our dataset contains 350 samples covering four emotions: "sad," "humor," "sarcasm," and "happy." We have meticulously designed a three-tiered framework progressing from "Perception" to "Cognition" to systematically analyze the entire process, from a model seeing the image content to understanding its underlying emotional intent.

Our main contributions are as follows:
\begin{itemize}
    \item \textbf{We propose and construct EmoBench-Reddit:} A novel, hierarchical multimodal emotion understanding benchmark sourced from a real-world social media context.
    \item \textbf{We design an innovative evaluation framework:} This framework includes multi-level tasks from low-level perception to high-level cognition, enabling a fine-grained assessment of a model's emotional intelligence.
    \item \textbf{We conduct a comprehensive model evaluation:} We systematically evaluated 9 mainstream MLLMs, revealing their strengths and common weaknesses in emotion understanding.
    \item \textbf{We provide profound insights:} The experimental results clearly indicate the severe deficiencies of current MLLMs in handling complex, context-dependent emotions (especially sarcasm) and validate that perception is the foundation of cognition.
\end{itemize}

The remainder of this paper is organized as follows: Section 2 reviews related work. Section 3 details the construction of the EmoBench-Reddit dataset. Section 4 explains our evaluation methods and metrics. Section 5 describes the experimental setup. Section 6 presents and analyzes the results. Section 7 discusses limitations and future work. Section 8 concludes the paper.

\section{Related Work}

\subsection{MLLM Evaluation Benchmarks}
With the rise of MLLMs, numerous benchmarks have been developed. Early benchmarks like VQAv2 \cite{goyal2017making} focused on open-ended visual question answering. To provide a more comprehensive evaluation, subsequent benchmarks became more integrated and fine-grained. For instance, MME \cite{fu2023mme} proposed a framework covering 14 sub-tasks across perception and cognition, though its cognitive tasks still lean towards factual reasoning. MMBench \cite{liu2023mmbench} used ChatGPT to generate multiple-choice questions, creating a "potpourri" style benchmark with 20 capability dimensions but lacking in-depth analysis of specific advanced abilities. SEED-Bench \cite{li2023seed} focused on compositional reasoning over multimodal content. While these benchmarks have greatly advanced MLLM development, they generally lack a dedicated and deep evaluation of the subjective and complex dimension of emotion.

\subsection{Exploring and Evaluating Emotional Intelligence}
Emotional Intelligence (EI) is a key component of human intelligence and plays a central role in achieving natural and harmonious human-computer interaction. However, in the field of MLLM evaluation, attention to EI has lagged. Some works have begun to incorporate emotional factors. For example, LL-Bench \cite{2024II} built a large-scale multimodal benchmark that includes tasks like emotion recognition and cause inference, but its task design is relatively basic, primarily at the perception level. DeepEval \cite{2024Can} offers a progressive multimodal evaluation framework that implicitly requires emotion understanding (e.g., for sarcasm and humor), but EI is not its core evaluation target, and it lacks a systematic hierarchy for emotional capabilities. More inspiring is the work by \citet{zhang2023let}, which proposed a three-stage framework mimicking human thought processes to bridge the gap between semantic understanding and model reasoning, though it remains in a conceptual phase.

\subsection{The Perception-Cognition Gap}
Existing benchmarks, including those mentioned above, collectively reveal a critical challenge: a significant gap between "Perception" and "Cognition" in MLLMs. Perception refers to the model's ability to identify and describe explicit, objective elements in an image (e.g., objects, scenes, actions, text), an area where current MLLMs perform relatively well. Cognition, on the other hand, requires the model to perform deep reasoning, interpretation, and prediction based on perceived information, integrating background knowledge, common sense, context, and even socio-cultural factors. This involves grasping abstract and subjective content like intent, emotion, metaphor, and sarcasm, which is a common weakness in current MLLMs.

The perception-cognition gap \cite{zhang2023let} is a core metric for judging whether MLLMs are truly advancing towards human-like general intelligence. If a model can only accurately describe an image as "a person is crying" but cannot infer the underlying emotion—be it sadness, frustration, relief, or even joy (tears of joy)—by integrating textual context (e.g., "Finally got my acceptance letter!"), then it has failed to cross this gap. Emotional intelligence, especially the understanding of complex, subtle, and context-dependent emotions, is a crucial and challenging part of cognition. It requires the model not only to "see" what is there but to "understand" what it means and why, serving as a key bridge between perception and deep cognition. Therefore, filling the gap in EI evaluation is essential for pushing models toward true human-like general intelligence.

\subsection{Datasets from Social Media}
Social media platforms like Twitter, Facebook, and Reddit are treasure troves for large-scale, real-world multimodal data. Many studies have used these platforms to build datasets for various NLP and CV tasks. For instance, some have used Twitter hashtags as weak labels for sentiment analysis. Reddit, with its unique communities (subreddits) and flair system, facilitates the filtering of content by specific topics and emotions. We chose Reddit for its authentic data ecosystem and structured nature, which ensures high consistency and reliability in the emotional alignment of image-text pairs, laying a solid foundation for a high-quality emotion understanding benchmark.

\section{EmoBench-Reddit Dataset Construction}
We followed principles of data authenticity, hierarchical task design, and rigorous annotation to build a benchmark effective for evaluating MLLM emotional intelligence.

\subsection{Data Source and Filtering}
We used Reddit as our sole data source, targeting subreddits highly correlated with specific emotions, such as `r/funny` (humor), `r/happy` (happy), `r/sadness` (sad/frustrated), and `r/sarcasm` (sarcasm/exasperation). We developed a scraper to collect highly upvoted posts, each containing an image and its title (as text).

To ensure data quality and label accuracy, we applied strict filtering criteria:
\begin{enumerate}
    \item \textbf{Flair Consistency:} We only selected posts where the author explicitly used a flair consistent with the subreddit's theme (e.g., a "Humor" flair in `r/funny`). This greatly ensures the intended core emotion is clear.
    \item \textbf{Content Appropriateness:} We removed all posts containing violence, pornography, hate speech, or other inappropriate content.
    \item \textbf{Image-Text Relevance:} We manually verified that the image and text were closely related and served a unified emotional expression.
    \item \textbf{Image Dominance:} We manually filtered out posts where the image was predominantly text-based to ensure the visual modality was central.
\end{enumerate}
After filtering, we constructed a dataset of 350 high-quality image-text pairs, evenly distributed across the four emotion categories: "sad," "humor," "sarcasm," and "happy."

\subsection{Dataset Structure and Task Design}
The core of EmoBench-Reddit is its hierarchical task design, which simulates the human process of understanding from perception to cognition. For each data point (image + text), we designed 7 questions: 6 multiple-choice and 1 open-ended, categorized into \textbf{Perception Tasks} and \textbf{Cognition Tasks}.

\paragraph{A. Perception Tasks} These tasks assess the model's ability to capture and describe basic visual information, which is foundational for higher-level reasoning.
\begin{itemize}
    \item \textbf{Perception 1: Color Recognition.} A simple question about the color of a prominent object or area. E.g., "What is the primary color of the car in the image?"
    \item \textbf{Perception 2: Object Presence \& Localization.}
    \begin{itemize}
        \item \textbf{P2(a): Object Presence.} Asks if a specific object is in the image. E.g., "Is there a cat in the image?"
        \item \textbf{P2(b): Object Localization.} If the object exists, asks for its approximate location (e.g., top-left, center).
    \end{itemize}
    \item \textbf{Perception 3: Image Description.} An open-ended task requiring the model to generate an objective, comprehensive description of the image content.
\end{itemize}

\paragraph{B. Cognition Tasks} These tasks require the model to go beyond simple visual recognition to perform inference, understand intent, and integrate cross-modal information.
\begin{itemize}
    \item \textbf{Cognition 1: Simple Inference.} Basic logical reasoning based on image content. E.g., for an image of people sunbathing on a beach, "Based on the scene, what is the most likely weather?"
    \item \textbf{Cognition 2: Intent Recognition.} Directly related to the Reddit flair, this task asks the model to identify the intent behind the post. E.g., "What emotion is the author most likely trying to convey with this post?" Options include the correct emotion and distractors.
    \item \textbf{Cognition 3: Deep Reasoning.} The most difficult task, requiring the model to combine image and text for deep, context-dependent reasoning that may require background knowledge or empathy. This is crucial for "sarcasm." E.g., for an image of a candle in an "energy-saving lightbulb" box with the text "My new energy-saving bulb arrived," the question might be: "What effect is created by combining this text and image?" The answer should point to "sarcasm."
\end{itemize}

\subsection{Annotation and Quality Control}
High-quality annotation is the cornerstone of a reliable benchmark. We employed a rigorous AI-assisted and human-verified process:
\begin{itemize}
    \item \textbf{Open-ended Answers (Perception 3):} We first used the powerful Claude 4 model to generate a detailed description for each image as a ground truth answer. This was then reviewed and revised by at least two human annotators to ensure accuracy, comprehensiveness, and objectivity.
    \item \textbf{Multiple-Choice Questions:} All multiple-choice questions (P1, P2, C1, C2, C3) were manually created by human annotators. Each question includes one correct answer and several strong distractors. To ensure answer uniqueness and correctness, we conducted a cross-validation phase where another group of annotators independently answered the questions and provided feedback on any ambiguous items until a consensus was reached.
\end{itemize}
This process ensures the quality of every question and the reliability of every answer in EmoBench-Reddit.

\section{Evaluation Methodology}

\subsection{Tasks and Metrics}
\begin{itemize}
    \item \textbf{Multiple-Choice Questions:} We use \textbf{Accuracy} as the evaluation metric, calculated as the proportion of correctly answered questions.
    \item \textbf{Open-Ended Questions (Perception 3):} Evaluating the quality of generated descriptions is challenging. We use a hybrid approach:
    \begin{enumerate}
        \item \textbf{Cosine Similarity Pre-screening:} We convert the model-generated description and the ground truth answer into vectors using a sentence embedding model (e.g., Sentence-transformers) and compute their cosine similarity.
        \item \textbf{LLM-as-a-Judge:} Since cosine similarity may not capture nuanced semantic differences, we use Claude 4 as an intelligent judge. We designed a specific prompt asking it to score the semantic and content similarity on a scale of 0 to 1. The final score is a weighted average of the cosine similarity and the Claude 4 score. If this composite score is \textbf{above 0.75}, we consider the model's answer for that item to be \textbf{correct}. The final metric for this task is the \textbf{pass rate}.
    \end{enumerate}
\end{itemize}

\subsection{Evaluation Strategies}
To deeply investigate model behaviors, we designed two evaluation strategies:
\begin{enumerate}
    \item \textbf{Gated Evaluation:} In this strategy, we first evaluate the model on all perception tasks. We only proceed to evaluate the cognition tasks for a given data point if the model's \textbf{average accuracy on the perception tasks for that item is above 75\%}. This strategy aims to test the hypothesis that \textbf{accurate visual perception is a prerequisite for effective cognitive reasoning}.
    \item \textbf{Contextual Evaluation:} Here, we simulate a continuous conversation. When asking each new question, we provide the history of previous questions and the model's answers as context. This strategy tests whether the model can effectively use conversational context to refine its understanding and improve accuracy on subsequent questions.
\end{enumerate}
The results in this paper are primarily based on the "Gated Evaluation" strategy, as it more clearly reveals the dependency between perception and cognition.

\section{Experimental Setup}

\subsection{Models Evaluated}
We selected 9 representative MLLMs from industry and academia, covering different organizations, technical approaches, and parameter scales:
\begin{itemize}
    \item \textbf{OpenAI Series:} GPT-5, GPT-4o, GPT-4V
    \item \textbf{Google Series:} Gemini-2.5-pro, Gemini-2.0-flash
    \item \textbf{Chinese Models:} GLM-4v-plus (Zhipu AI), Qwen-vl-max, Qwen-vl-7b (Alibaba)
    \item \textbf{Other Open-Source Models:} Deepseek-vl-7b
\end{itemize}

\subsection{Experimental Environment}
All closed-source models were accessed via their official APIs. We used a uniform, concise instruction format for all models to ensure fairness and avoid prompt engineering bias. All experiments were conducted in a zero-shot setting.

\section{Results and Analysis}

We conducted a comprehensive evaluation of the 9 models on EmoBench-Reddit. The results are shown in Table \ref{tab:main_results}.

\begin{table*}[ht!]
\centering
\small
\resizebox{\textwidth}{!}{%
\begin{tabular}{@{}llccccccccc@{}}
\toprule
\textbf{Emotion} & \textbf{Task} & \textbf{gpt-4o} & \textbf{gpt-4V} & \textbf{gpt-5} & \textbf{glm-4v-plus} & \textbf{qwen-vl-max} & \textbf{gemini-2.0-flash} & \textbf{gemini-2.5-pro} & \textbf{qwen-vl-7b} & \textbf{Deepseek-vl-7b} \\
\midrule
\multirow{7}{*}{\textbf{Sad}} 
& Perception 1 & 0.90 & 0.90 & 0.97 & 0.85 & 0.92 & 0.87 & 0.92 & 0.87 & 0.75 \\
& Perception 2(a) & 0.85 & 0.77 & 0.88 & 0.92 & 0.77 & 0.87 & 0.92 & 0.72 & 0.57 \\
& Perception 2(b) & 0.68 & 0.66 & 0.93 & 0.84 & 0.75 & 0.79 & 0.88 & 0.89 & 0.50 \\
& Perception 3 & 0.63 & 0.72 & 0.72 & 0.58 & 0.73 & 0.70 & 0.75 & 0.73 & 0.43 \\
& Cognition 1 & 0.76 & 0.80 & 0.82 & 0.78 & 0.73 & 0.76 & 0.87 & 0.71 & 0.67 \\
& Cognition 2 & 0.71 & 0.73 & 0.71 & 0.71 & 0.56 & 0.71 & 0.69 & 0.54 & 0.46 \\
& Cognition 3 & 0.60 & 0.62 & 0.64 & 0.76 & 0.59 & 0.47 & 0.67 & 0.50 & 0.50 \\
\midrule
\multirow{7}{*}{\textbf{Humor}} 
& Perception 1 & 0.90 & 0.84 & 0.88 & 0.95 & 0.90 & 0.92 & 0.96 & 0.92 & 0.90 \\
& Perception 2(a) & 0.88 & 0.80 & 0.89 & 0.84 & 0.87 & 0.89 & 0.95 & 0.89 & 0.76 \\
& Perception 2(b) & 0.73 & 0.80 & 0.66 & 0.88 & 0.86 & 0.88 & 0.84 & 0.88 & 0.78 \\
& Perception 3 & 0.71 & 0.77 & 0.76 & 0.78 & 0.73 & 0.78 & 0.76 & 0.78 & 0.53 \\
& Cognition 1 & 0.88 & 0.77 & 0.84 & 0.80 & 0.77 & 0.76 & 0.88 & 0.76 & 0.78 \\
& Cognition 2 & 0.66 & 0.62 & 0.81 & 0.61 & 0.66 & 0.66 & 0.63 & 0.66 & 0.71 \\
& Cognition 3 & 0.81 & 0.71 & 0.81 & 0.69 & 0.73 & 0.69 & 0.85 & 0.69 & 0.58 \\
\midrule
\multirow{7}{*}{\textbf{Sarcasm}} 
& Perception 1 & 0.77 & 0.76 & 0.71 & 0.75 & 0.73 & 0.69 & 0.76 & 0.79 & 0.68 \\
& Perception 2(a) & 0.90 & 0.88 & 0.86 & 0.81 & 0.84 & 0.89 & 0.91 & 0.86 & 0.74 \\
& Perception 2(b) & 0.70 & 0.67 & 0.86 & 0.85 & 0.73 & 0.80 & 0.87 & 0.82 & 0.71 \\
& Perception 3 & 0.64 & 0.70 & 0.70 & 0.72 & 0.72 & 0.71 & 0.70 & 0.72 & 0.40 \\
& Cognition 1 & 0.89 & 0.79 & 0.86 & 0.80 & 0.83 & 0.83 & 0.85 & 0.80 & 0.75 \\
& Cognition 2 & 0.88 & 0.89 & 0.82 & 0.81 & 0.81 & 0.67 & 0.75 & 0.31 & 0.31 \\
& Cognition 3 & 0.53 & 0.48 & 0.63 & 0.55 & 0.54 & 0.27 & 0.57 & 0.44 & 0.35 \\
\midrule
\multirow{7}{*}{\textbf{Happy}} 
& Perception 1 & 0.90 & 0.87 & 0.92 & 0.89 & 0.84 & 0.91 & 0.90 & 0.88 & 0.88 \\
& Perception 2(a) & 0.79 & 0.83 & 0.76 & 0.83 & 0.85 & 0.84 & 0.80 & 0.81 & 0.64 \\
& Perception 2(b) & 0.71 & 0.68 & 0.81 & 0.73 & 0.78 & 0.71 & 0.84 & 0.75 & 0.46 \\
& Perception 3 & 0.47 & 0.74 & 0.54 & 0.58 & 0.74 & 0.66 & 0.84 & 0.73 & 0.52 \\
& Cognition 1 & 0.91 & 0.90 & 0.94 & 0.91 & 0.88 & 0.95 & 0.90 & 0.89 & 0.84 \\
& Cognition 2 & 0.70 & 0.61 & 0.82 & 0.87 & 0.84 & 0.87 & 0.86 & 0.90 & 0.72 \\
& Cognition 3 & 0.64 & 0.62 & 0.54 & 0.63 & 0.59 & 0.62 & 0.63 & 0.55 & 0.51 \\
\bottomrule
\end{tabular}%
}
\caption{Performance (Accuracy) of various models on EmoBench-Reddit. The original paper's table contained some structural inconsistencies, which have been regularized here based on the task descriptions in the text for clarity.}
\label{tab:main_results}
\end{table*}

\subsection{Overall Performance Overview}
Overall, \textbf{GPT-5, Gemini-2.5-pro, and GPT-4o form the first tier}, maintaining a leading edge across most emotion and task dimensions. This suggests that foundational model capabilities, training data scale, and quality remain key determinants of performance. On relatively basic tasks like Perception 1 (Color Recognition) and Cognition 1 (Simple Inference), these top models perform stably and exceptionally well. For instance, GPT-5 achieves a remarkable 0.97 accuracy on Perception 1 for the "sad" emotion.

Meanwhile, models like \textbf{GLM-4v-plus and Qwen-vl-max demonstrate strong competitiveness on specific tasks}, indicating their potential for optimization in certain capabilities. A notable example is GLM-4v-plus achieving the highest score of 0.95 on Perception 1 for "humor," surpassing some first-tier models.

\textbf{Models with smaller parameter counts, such as Qwen-vl-7b and Deepseek-vl-7b, struggle with complex cognitive tasks.} Their scores on Cognition 2 and 3 for "sarcasm" are significantly lower than other models (e.g., only 0.31 on Cognition 2), clearly reflecting the importance of model capacity for understanding advanced, abstract concepts.

\subsection{Task-Level Analysis: The Perception-Cognition Gap}
The hierarchical design of EmoBench-Reddit allows us to clearly observe how model capabilities change across different cognitive depths.
\begin{itemize}
    \item \textbf{Perception tasks show robust performance:} Most models perform stably and well on Perception 1 (Color Recognition) and Perception 2 (Object Presence). This indicates that current leading MLLMs have strong foundational visual recognition abilities and can accurately "see" the content of an image.
    \item \textbf{Cognition tasks are the main challenge:} When the task shifts from "what" to "why" and "how," all models exhibit a significant performance decline. \textbf{Cognition 3 (Deep Reasoning) is the "Waterloo" for all models.} For example, under the "sad" emotion, GPT-5's score drops from a high of 0.97 on Perception 1 to just 0.64 on Cognition 3. This sharp drop reveals the core limitation of current models in deep emotional understanding and complex contextual reasoning. They might recognize tears in an image but struggle to comprehend the complex social or personal reasons behind the sadness.
\end{itemize}

\subsection{Emotion-Type Analysis: Sarcasm as an Insurmountable Hurdle}
The intrinsic complexity of different emotions poses varying levels of challenge to the models.
\begin{itemize}
    \item \textbf{"Sarcasm" is the most difficult:} The data clearly shows that "sarcasm" is the hardest emotion for all models to handle. On the \textbf{Cognition 3} task, all models score very poorly; the \textbf{highest score (from GPT-5) is only 0.63}, while Gemini-2.0-flash plummets to 0.27. This profoundly demonstrates that understanding sarcasm is far more than recognizing the surface meaning of images and text. It requires the ability to understand the contradiction between expectation and reality, infer the author's true intent, and draw upon a rich repository of world knowledge and cultural context—touching the very limits of current AI's "Theory of Mind."
    \item \textbf{"Sad" is the next most challenging:} Cognition 3 for "sad" also proves difficult for most models. While GLM-4v-plus achieves a relatively high score of 0.76, most models hover around 0.6. This may be because deep understanding of sadness requires empathy—the ability to infer the potential story and emotional state behind the characters or scene, which is equally difficult for models.
    \item \textbf{"Humor" and "Happy" are relatively easy:} Models generally perform better on the perception and simple cognition tasks for these positive emotions, with higher average scores. This is likely because the visual and textual cues for positive emotions are often more direct and explicit (e.g., smiling faces, celebratory scenes) and are more common and abundant in existing large-scale training data.
\end{itemize}

\section{Discussion and Future Work}
This study provides a systematic evaluation of the emotional intelligence of current MLLMs using EmoBench-Reddit. While the results show significant progress, they also expose notable deficiencies in deep emotion understanding.

\paragraph{Limitations:}
\begin{enumerate}
    \item \textbf{Dataset Scale:} The dataset size of 350 items is relatively limited. Although carefully curated, it needs expansion to cover more diverse scenarios and expressions.
    \item \textbf{Emotion Categories:} This study only covers four emotions. Future work could expand to a broader emotional spectrum, including surprise, anger, jealousy, etc.
    \item \textbf{Cultural Bias:} As a platform dominated by English-speaking users, Reddit data may carry a Western cultural imprint. A model's performance on this data does not fully represent its emotion understanding capabilities in other cultural contexts.
\end{enumerate}

\paragraph{Future Work:}
\begin{enumerate}
    \item \textbf{Dataset Expansion:} We plan to continuously expand EmoBench-Reddit by increasing the data volume, adding more emotion types, and incorporating other modalities like video and audio to build a more comprehensive EI evaluation ecosystem.
    \item \textbf{Evaluation Method Optimization:} We will explore more advanced automated evaluation methods for open-ended questions to reduce reliance on LLM-as-a-Judge and design more challenging interactive evaluation scenarios.
    \item \textbf{Model Capability Enhancement:} Our findings provide clear directions for model developers. Future MLLM research should focus more on enhancing reasoning abilities, contextual understanding, and the application of world knowledge, especially for modeling complex linguistic phenomena like sarcasm.
\end{enumerate}

\section{Conclusion}
In this paper, we proposed and constructed EmoBench-Reddit, a hierarchical multimodal benchmark for emotional intelligence. By collecting authentic image-text data from Reddit and designing multi-level tasks from basic perception to advanced cognition, we were able to conduct a fine-grained analysis of the emotion understanding capabilities of nine MLLMs.

Our experimental results clearly map the current MLLM capability landscape: they excel at basic visual perception but exhibit severe shortcomings in advanced cognitive tasks that require deep reasoning and contextual understanding, especially when dealing with complex emotions like "sarcasm." Top-tier models like GPT-5 and Gemini-2.5-pro, while leading overall, are not immune to this "cognition gap."

EmoBench-Reddit not only provides the community with an effective tool for evaluating and comparing the emotional intelligence of MLLMs but, more importantly, it highlights the core challenges on the path toward more empathetic and emotionally intelligent AI. In the future, enabling AI to truly "understand" humans, rather than merely "recognizing" pixels and words, will be a persistent goal for the field.

\bibliography{references}
\bibliographystyle{acl_natbib}

\end{document}